\documentclass[runningheads]{llncs}

\usepackage{adjustbox}

\usepackage{subcaption} % For subfigures

\usepackage{multirow}

\usepackage[T1]{fontenc}
\usepackage{graphicx}
\begin{document}
\title{
%Deep Periocular Recognition via Ocular Cropping from Large-Scale Face Datasets
Leveraging Large-Scale Face Datasets for Deep Periocular Recognition via Ocular Cropping
%\thanks{Supported by organization x.}
}
\author{Fernando Alonso-Fernandez\inst{1} \and
Kevin Hernandez-Diaz\inst{1} \and \\
Jose Maria Buades Rubio\inst{2} \and
Josef Bigun\inst{1}}
\authorrunning{F. Alonso-Fernandez et al.}
\institute{School of Information Technology, Halmstad University, Sweden \\ \email{feralo@hh.se, kevin.hernandez-diaz@hh.se, josef.bigun@hh.se} \and Computer Graphics and Vision and AI Group, University of Balearic Islands, Spain \email{josemaria.buades@uib.es}}
\maketitle              % typeset the header of the contribution
\begin{abstract}

We focus on ocular biometrics, specifically the periocular region (the area around the eye), which offers high discrimination and minimal acquisition constraints.
We evaluate three Convolutional Neural Network architectures of varying depth and complexity %(SqueezeNet, MobileNetv2, and ResNet50), 
to assess their effectiveness for periocular recognition.
The networks are trained on 1,907,572 ocular crops extracted from the large-scale
VGGFace2 database.
This significantly contrasts with existing works, which typically rely on small-scale periocular datasets for training having only a few thousand images.
Experiments are conducted with ocular images from VGGFace2-Pose,
a subset of VGGFace2 containing in-the-wild face images, and the UFPR-Periocular database, which consists of selfies captured via mobile devices with user guidance on the screen.
Due to the uncontrolled conditions of VGGFace2, the Equal Error Rates (EERs) obtained with ocular crops range from 9–15\%, noticeably higher than the 3–6\% EERs achieved using full-face images. 
In contrast, UFPR-Periocular yields significantly better performance (EERs of 1–2\%), thanks to higher image quality and more consistent acquisition protocols. 
To the best of our knowledge, these are the lowest reported EERs on the UFPR dataset to date.

\end{abstract}

\section{Introduction}

%\section{Background}
%\label{sect:background}

The periocular region (the area surrounding the eye) offers a robust alternative to face and iris modalities, especially under challenging conditions such as occlusion, low resolution, or poor imaging, situations where even basic face or iris detection may fail \cite{Alonso24computers_periSOA}.
Partial faces can also be an issue in controlled contexts such as social media \cite{Hedman22_pr_selfie_beauty_filters}, masks, professional work gear, cultural coverings, etc. \cite{sharma23cviu_periocular_masks_survey}.
In this regard, periocular recognition has rapidly emerged as a promising
approach for unconstrained biometrics \cite{Alonso24computers_periSOA,sharma23cviu_periocular_masks_survey,Kumari22_jksu_periocular_survey,[Rattani17soaOcularVIS],[Alonso16]}.
As with many other vision tasks, Convolutional Neural Networks (CNNs) have gained popularity in biometrics \cite{[Sundararajan18-DLbiometrics]}. 
However, their application to periocular remains limited \cite{sharma23cviu_periocular_masks_survey,zanlorensi22_AIR_ocular_db_competitions_survey,zeng21_iet_FR_occlusion_survey}, 
primarily due to the scarcity of large databases \cite{zanlorensi22_AIR_ocular_db_competitions_survey}.

Recent works \cite{zanlorensi2022_SR_UFPR_db,Kolf22ijcb_light_ocular_lowbit_quantization,Rattani23ACCESS_OcularCNNPruningBenchmark,Kolf23_ivc_syper_ocular_db,Kolf24_eaai_MixQuantBio_face_ocular,Coelho24_lacci_PeriocularEfficientNet,Carreira24_SIBGRAPI_deep_ocular_surveillance} primarily relied on small to medium-scale datasets for periocular recognition training, such as UFPR-Periocular \cite{zanlorensi2022_SR_UFPR_db} (33,660 ocular images) and VISOB 2.0 \cite{Nguyen21_icip_visob2} (158,136 images).
This contrasts with face recognition research, which benefits from datasets with millions of images \cite{Zhu23_pami_WebFace260M}.
For instance, \cite{zanlorensi2022_SR_UFPR_db} benchmarked seven CNN architectures trained on the UFPR database, initialized with either ImageNet or face recognition weights.
Studies in  \cite{Kolf22ijcb_light_ocular_lowbit_quantization,Kolf23_ivc_syper_ocular_db,Kolf24_eaai_MixQuantBio_face_ocular} proposed lightweight periocular architectures via quantization techniques, using UFPR to train models such as ResNet18, ResNet50, and MobileFaceNet from scratch.
Additionally, the authors of \cite{Kolf23_ivc_syper_ocular_db} trained a Generative Adversarial Network (GAN) to generate 99,840 synthetic periocular images, which were added to the training set. 
In \cite{Rattani23ACCESS_OcularCNNPruningBenchmark}, several network compression techniques were evaluated in the context of ocular recognition using five CNN models based on ResNet and VGG architectures. As training sets, the authors employed UFPR and VISOB 2.0 from scratch.
The work \cite{Coelho24_lacci_PeriocularEfficientNet} adapted the EfficientNet architecture for periocular recognition using UFPR, starting from the ImageNet trained model.
Another recent work \cite{Carreira24_SIBGRAPI_deep_ocular_surveillance} adopted a strategy similar to ours, using ocular crops from VGGFace2 \cite{[Cao18vggface2]} to train a face-pretrained ResNet50. 
The paper, which does not specify the amount of images gathered for training, used the Cox database for evaluation, which contains surveillance videos.

In the present work, we explore deep periocular recognition using large-scale face datasets, addressing the limitations posed by the scarcity of dedicated ocular databases.
Specifically, we train three convolutional networks (SqueezeNet, MobileNetv2, and ResNet50) using over 1.9 million ocular crops extracted from the VGGFace2 dataset.
Different to \cite{Carreira24_SIBGRAPI_deep_ocular_surveillance}, we evaluate here multiple network initializations and architectures, as well as their fusion.
We evaluate the trained models on two benchmarks: VGGFace2-Pose, containing unconstrained in-the-wild images, and UFPR-Periocular, a more controlled selfie dataset captured at close distance by guiding users to align their eyes within a region shown on the screen. 
Given such difference in image quality, EERs with VGGFace2-Pose are modest (9-15\%) compared to UFPR (1-2\%).
To the best of our knowledge, the results reported here are the best published EERs on the UFPR dataset to date.

%zanlorensi2022_SR_UFPR_db,

%Kolf22ijcb_light_ocular_lowbit_quantization,

%Rattani23ACCESS_OcularCNNPruningBenchmark,

%Kolf23_ivc_syper_ocular_db,

%Kolf24_eaai_MixQuantBio_face_ocular,

%Coelho24_lacci_PeriocularEfficientNet

\section{Materials and Methods}

\subsection{Recognition Networks}
\label{sect:networks}

We use three backbone architectures:  SqueezeNet \cite{[Iandola16SqueezeNet]} (light), MobileNetv2 \cite{[Sandler18mobilenetv2]} (medium) and ResNet50 \cite{[He16]} (large). 
They respectively have 18/53/50 convolutional layers and 1.24M/3.5M/25.6M parameters.
ResNet introduced residual blocks that bypass intermediate layers, improving gradient propagation and allowing deeper networks without overfitting. 
In a residual layer, channel dimensionality is first reduced via 1$\times$1 point-wise filters, after which larger 3$\times$3 filters are applied in a reduced space,  
%To enable a residual connection, channel dimensionality is then increased to match the input via 1$\times$1 filters.
%
to have dimensionality increased again to match the input.
MobileNets employ inverted residuals and depth-wise separable convolutions to reduce parameters and inference time.
Shortcut connections are between thinner layers instead (hence the name 'inverted'), which also results in fewer parameters, whereas the intermediate representation lies in a higher dimensional space. 
%
%To reduce the number of parameters and the inference time, MobileNets employ inverted residuals, in which shortcut connections are between thinner layers instead (hence the name 'inverted'). 
%%The input and output lie in a reduced dimensional space, whereas the intermediate representation lies in a higher dimensional space. 
%
%Also, 3$\times$3 filters are applied via depth-wise separable convolutions, which have fewer parameters and are faster than regular filters.
%
%Thanks to the mentioned techniques, MobileNets contain the same amount of convolutional layers ($\sim$50 in our case) with significantly less parameters.
%
SqueezeNet, on the other hand, is a sequential (non-residual) network which is among the smallest generic CNNs proposed in the context of ImageNet. %, and one of the early networks designed to reduce the number of parameters and size.
It also applies the 1$\times$1 point-wise convolution paradigm to reduce (\textit{squeeze}) the channel dimensionality and then apply a larger amount of (more costly) 3$\times$3 and 1$\times$1 filters in a lower dimensional space (\textit{expand} phase).

This choice allows comparison of networks of different sizes.
%
%\textbf{Table XX shows...}
%
We use the models loaded in our experimental environment (Matlab r2024b), modified to have an input of 113$\times$113 by changing the stride of the first convolutional layer from 2 to 1.
This allows to keep the network unchanged and reuse ImageNet as starting weights when appropriate.
Input images are normalised by subtracting 127.5 and dividing by 128.
For SqueezeNet, we adopt modifications of \cite{[Alonso20SqueezeFacePoseNet]}, which added batch norm between convolutions and ReLU (missing in the original model).

\begin{figure}[t]
\centering
\includegraphics[width=0.88\textwidth]{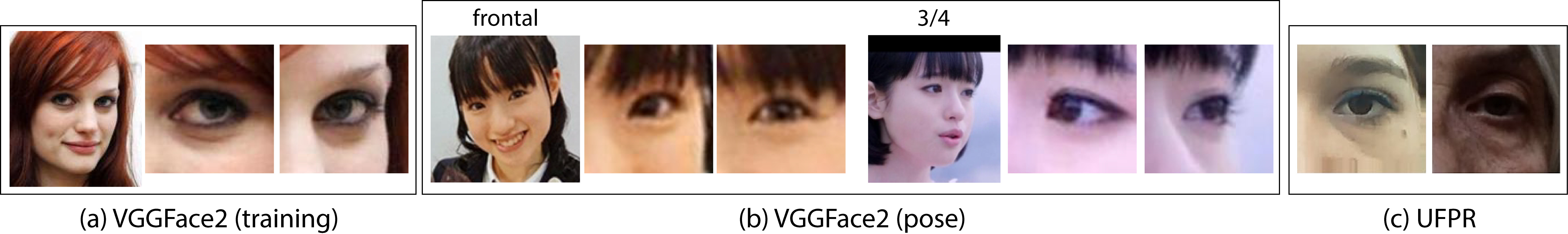}
\caption{Example images of the databases employed. Image c) composed from \cite{zanlorensi2022_SR_UFPR_db}.} \label{fig:databases}
\end{figure}

\subsection{Databases}
\label{sect:database}

%This sub-section describes the databases employed (Figure~\ref{fig:databases}).
%
We use VGGFace2 (3.31M images, 9131 identities) \cite{[Cao18vggface2]} for training and evaluation (Figure~\ref{fig:databases}). 
The dataset includes significant variation in pose, age, lighting, and background. 
We use the database annotation to crop the ocular regions. 
The training protocol considers 8631 training classes (3.14M images). 
Images are aligned (eye centres horizontal), scaled to 113 pixels inter-eye distance, and cropped into two 113$\times$113 patches centred on each eye. 
We apply a loose frontality check to ensure that both eyes are visible, imposing that the distance between the centre of the eyes and the vertical of the nose must be below 40\% of the inter-eye distance. 
Faces with original inter-eye distances $<$50 px are also discarded to avoid excessive upsampling. 
Left eye crops are flipped for orientation consistency, and both eyes are then treated as the same identity. 
This results in 953,786 valid faces and 1,907,572 ocular crops (221 per identity on average).

For testing, we use the VGGFace2-Pose subset, with 368 subjects and 10 images per pose (frontal, three-quarter, profile). 
Only frontal and three-quarter are used, since profiles likely miss one of the eyes, and the available one can be severely distorted. This yields 7,360 crops per pose (20 per subject). 
To account for distortion, three-quarter images are resized to 80 pixels inter-eye distance instead.
We also evaluate on UFPR-Periocular \cite{zanlorensi2022_SR_UFPR_db}, the latest and one of the largest ocular databases, with 33,660 eye images from 1,122 subjects across 3 sessions using 196 mobile devices.
Images vary in blur, occlusion, and lighting to simulate real-world conditions. 
UFPR contains three different training and evaluation protocols.
We follow the open world/closed validation (OW/CW), where test identities are not included in the training/validation set.
Eye crops (224$\times$224 pixels) are provided, which we resize to 113$\times$113 to fit the input of the CNNs. 
We also flip left eyes, treating both eyes as the same identity.

In some configurations, networks are first pretrained for face recognition. Following \cite{[Alonso20SqueezeFacePoseNet]}, ImageNet initialised models are trained on the RetinaFace cleaned MS1M dataset \cite{[Guo16_MSCeleb1M]} (5.1M face images, 93.4K identities, 113$\times$113 pixels).
Then, they are fine-tuned on VGG2 (3.14M face images). 
This two-step face training approach has demonstrated superior performance \cite{[Cao18vggface2],[Alonso20SqueezeFacePoseNet]}, leveraging the large image count of MS1M and the greater intra-class diversity of VGG2 due to having more images per person.

\begin{table}[t]
\centering
\caption{Verification scores with VGGFace2-Pose and UFPR.}

\label{tab:scores}

\begin{adjustbox}{max width=0.95\textwidth}

%\scalebox{0.7}{

%\small
%\footnotesize
%\scriptsize
%\begin{center}
\begin{tabular}{|c|c|c|c|c|c|c|c|}

\multicolumn{8}{c}{} \\ \cline{1-2} \cline{4-5} \cline{7-8}

\multicolumn{2}{|c|}{\textbf{VGG SAME-POSE}}  & \multicolumn{1}{c}{} & \multicolumn{2}{|c|}{\textbf{VGG CROSS-POSE}} & \multicolumn{1}{c}{} & \multicolumn{2}{|c|}{\textbf{UFPR (per fold)}} \\   \cline{1-2} \cline{4-5} \cline{7-8}

genuine & impostor & \multicolumn{1}{c|}{} & genuine & impostor & \multicolumn{1}{c|}{} & genuine & impostor \\  \cline{1-2} \cline{4-5} \cline{7-8}

368$\times$(9+8+...+1)=16560  & 368$\times$36 =135056 & \multicolumn{1}{c|}{} & 368$\times$10$\times$10 = 36800 &  368$\times$367=135056  &  & 78,540 & 4,190,670 \\  \cline{1-2} \cline{4-5} \cline{7-8}

%\multicolumn{8}{c}{} \\

\end{tabular}

%\end{center}

%}

\end{adjustbox}

\end{table}
%\normalsize

% \usepackage{multirow}
\begin{table}[t]
\centering
\caption{Ocular verification results on VGGFace2-Pose for different network intializations (EER \%). The best result of each network (per column) is in bold. The table also shows full-face results from previous works on the same database.}

\label{tab:eer_vgg2_pose}

\begin{adjustbox}{max width=0.85\textwidth}

\begin{tabular}{llcccccccc}
 &  & \multicolumn{1}{l}{} & \multicolumn{1}{l}{} & \multicolumn{1}{l}{} & \multicolumn{1}{l}{} & \multicolumn{1}{l}{} & \multicolumn{1}{l}{} & \multicolumn{1}{l}{} & \multicolumn{1}{l}{} \\ \cline{3-10} 
 
 & \multicolumn{1}{l||}{} & \multicolumn{4}{c||}{\textbf{cosine similarity}} & \multicolumn{4}{c||}{\textbf{$\chi^2$ distance}} \\ \hline \hline
 
\multicolumn{1}{|l|}{\textbf{Net}} & \multicolumn{1}{l||}{\textbf{Initialization}} & \multicolumn{1}{c|}{\textbf{frontal}} & \multicolumn{1}{c|}{\textbf{3/4}} & \multicolumn{1}{c|}{\textbf{cross}} & \multicolumn{1}{c||}{\textbf{all}} & \multicolumn{1}{c|}{\textbf{frontal}} & \multicolumn{1}{c|}{\textbf{3/4}} & \multicolumn{1}{c|}{\textbf{cross}} & \multicolumn{1}{c||}{\textbf{all}} \\ \hline \hline

\multicolumn{1}{|l|}{\multirow{3}{*}{SQ}} & \multicolumn{1}{l||}{Scratch} & \multicolumn{1}{c|}{15.02} & \multicolumn{1}{c|}{15.87} & \multicolumn{1}{c|}{15.86} & \multicolumn{1}{c||}{15.79} & \multicolumn{1}{c|}{13.80} & \multicolumn{1}{c|}{\textbf{15.31}} & \multicolumn{1}{c|}{14.97} & \multicolumn{1}{c||}{14.96} \\ \cline{2-10} 
\multicolumn{1}{|l|}{} & \multicolumn{1}{l||}{ImageNet} & \multicolumn{1}{c|}{14.21} & \multicolumn{1}{c|}{\textbf{15.46}} & \multicolumn{1}{c|}{15.23} & \multicolumn{1}{c||}{15.15} & \multicolumn{1}{c|}{13.80} & \multicolumn{1}{c|}{\textbf{15.31}} & \multicolumn{1}{c|}{14.97} & \multicolumn{1}{c||}{14.96} \\ \cline{2-10} 
\multicolumn{1}{|l|}{} & \multicolumn{1}{l||}{Face} & \multicolumn{1}{c|}{\textbf{13.47}} & \multicolumn{1}{c|}{15.62} & \multicolumn{1}{c|}{\textbf{15.04}} & \multicolumn{1}{c||}{\textbf{14.95}} & \multicolumn{1}{c|}{\textbf{13.07}} & \multicolumn{1}{c|}{15.67} & \multicolumn{1}{c|}{\textbf{14.69}} & \multicolumn{1}{c||}{\textbf{14.80}} \\ \hline \hline

\multicolumn{1}{|l|}{\multirow{3}{*}{MB2}} & \multicolumn{1}{l||}{Scratch} & \multicolumn{1}{c|}{10.59} & \multicolumn{1}{c|}{12.05} & \multicolumn{1}{c|}{11.80} & \multicolumn{1}{c||}{11.66} & \multicolumn{1}{c|}{10.66} & \multicolumn{1}{c|}{12.07} & \multicolumn{1}{c|}{11.76} & \multicolumn{1}{c||}{11.66} \\ \cline{2-10} 
\multicolumn{1}{|l|}{} & \multicolumn{1}{l||}{ImageNet} & \multicolumn{1}{c|}{\textbf{8.93}} & \multicolumn{1}{c|}{\textbf{10.85}} & \multicolumn{1}{c|}{\textbf{10.23}} & \multicolumn{1}{c||}{\textbf{10.13}} & \multicolumn{1}{c|}{\textbf{8.56}} & \multicolumn{1}{c|}{\textbf{10.35}} & \multicolumn{1}{c|}{\textbf{9.72}} & \multicolumn{1}{c||}{\textbf{9.70}} \\ \cline{2-10} 
\multicolumn{1}{|l|}{} & \multicolumn{1}{l||}{Face} & \multicolumn{1}{c|}{9.74} & \multicolumn{1}{c|}{11.59} & \multicolumn{1}{c|}{11.02} & \multicolumn{1}{c||}{10.94} & \multicolumn{1}{c|}{9.77} & \multicolumn{1}{c|}{11.76} & \multicolumn{1}{c|}{11.09} & \multicolumn{1}{c||}{11.00} \\ \hline \hline

\multicolumn{1}{|l|}{\multirow{3}{*}{R50}} & \multicolumn{1}{l||}{Scratch} & \multicolumn{1}{c|}{\textbf{9.09}} & \multicolumn{1}{c|}{\textbf{10.10}} & \multicolumn{1}{c|}{\textbf{10.00}} & \multicolumn{1}{c||}{\textbf{9.85}} & \multicolumn{1}{c|}{\textbf{8.66}} & \multicolumn{1}{c|}{\textbf{9.71}} & \multicolumn{1}{c|}{\textbf{9.53}} & \multicolumn{1}{c||}{\textbf{9.41}} \\ \cline{2-10} 
\multicolumn{1}{|l|}{} & \multicolumn{1}{l||}{ImageNet} & \multicolumn{1}{c|}{9.80} & \multicolumn{1}{c|}{11.14} & \multicolumn{1}{c|}{10.79} & \multicolumn{1}{c||}{10.68} & \multicolumn{1}{c|}{8.75} & \multicolumn{1}{c|}{10.10} & \multicolumn{1}{c|}{9.68} & \multicolumn{1}{c||}{9.62} \\ \cline{2-10} 
\multicolumn{1}{|l|}{} & \multicolumn{1}{l||}{Face} & \multicolumn{1}{c|}{10.46} & \multicolumn{1}{c|}{12.05} & \multicolumn{1}{c|}{11.87} & \multicolumn{1}{c||}{11.66} & \multicolumn{1}{c|}{10.40} & \multicolumn{1}{c|}{12.02} & \multicolumn{1}{c|}{11.75} & \multicolumn{1}{c||}{11.56} \\ \hline \hline

\multicolumn{2}{|l||}{MB2+R50 (best init)} & \multicolumn{1}{c|}{-} & \multicolumn{1}{c|}{-} & \multicolumn{1}{c|}{-} & \multicolumn{1}{c||}{-} & \multicolumn{1}{c|}{\textbf{7.99}} & \multicolumn{1}{c|}{\textbf{9.27}} & \multicolumn{1}{c|}{\textbf{8.85}} & \multicolumn{1}{c||}{\textbf{8.83}} \\ \hline

\multicolumn{2}{|l||}{MB2+R50 (ImageNet)} & \multicolumn{1}{c|}{-} & \multicolumn{1}{c|}{-} & \multicolumn{1}{c|}{-} & \multicolumn{1}{c||}{-} & \multicolumn{1}{c|}{8.10} & \multicolumn{1}{c|}{9.63} & \multicolumn{1}{c|}{9.07} & \multicolumn{1}{c||}{9.04} \\ \hline 

\multicolumn{10}{l}{} \\ 

\multicolumn{10}{c}{\textbf{Face recognition} performance in another works of the literature} \\ \hline

\multicolumn{2}{|l||}{SqueezeNet \cite{[Alonso20SqueezeFacePoseNet]}} & \multicolumn{1}{c|}{-} & \multicolumn{1}{c|}{-} & \multicolumn{1}{c|}{-} & \multicolumn{1}{c||}{-} & \multicolumn{1}{c|}{6.39} & \multicolumn{1}{c|}{5.47} & \multicolumn{1}{c|}{6.09} & \multicolumn{1}{c||}{-} \\ \hline

\multicolumn{2}{|l||}{ResNet50ft \cite{[Alonso20SqueezeFacePoseNet]}} & \multicolumn{1}{c|}{-} & \multicolumn{1}{c|}{-} & \multicolumn{1}{c|}{-} & \multicolumn{1}{c||}{-} & \multicolumn{1}{c|}{4.14} & \multicolumn{1}{c|}{3.13} & \multicolumn{1}{c|}{3.68} & \multicolumn{1}{c||}{-} \\ \hline

\multicolumn{2}{|l||}{SENet50ft \cite{[Alonso20SqueezeFacePoseNet]}} & \multicolumn{1}{c|}{-} & \multicolumn{1}{c|}{-} & \multicolumn{1}{c|}{-} & \multicolumn{1}{c||}{-} & \multicolumn{1}{c|}{3.86} & \multicolumn{1}{c|}{2.87} & \multicolumn{1}{c|}{3.36} & \multicolumn{1}{c||}{-} \\ \hline \hline

\multicolumn{2}{|l||}{MobileNetv2 \cite{Alonso23wifs_lime_biometrics}} & \multicolumn{1}{c|}{3.69} & \multicolumn{1}{c|}{2.91} & \multicolumn{1}{c|}{3.33} & \multicolumn{1}{c||}{-} & \multicolumn{1}{c|}{-} & \multicolumn{1}{c|}{-} & \multicolumn{1}{c|}{-} & \multicolumn{1}{c||}{-} \\ \hline

\multicolumn{2}{|l||}{ResNet50 \cite{Alonso23wifs_lime_biometrics}} & \multicolumn{1}{c|}{3.93} & \multicolumn{1}{c|}{3.01} & \multicolumn{1}{c|}{3.51} & \multicolumn{1}{c||}{-} & \multicolumn{1}{c|}{-} & \multicolumn{1}{c|}{-} & \multicolumn{1}{c|}{-} & \multicolumn{1}{c||}{-} \\ \hline

\multicolumn{2}{|l||}{MB2+R50 \cite{Alonso23wifs_lime_biometrics}} & \multicolumn{1}{c|}{3.53} & \multicolumn{1}{c|}{2.70} & \multicolumn{1}{c|}{3.13} & \multicolumn{1}{c||}{-} & \multicolumn{1}{c|}{-} & \multicolumn{1}{c|}{-} & \multicolumn{1}{c|}{-} & \multicolumn{1}{c||}{-} \\ \hline

\end{tabular}

\end{adjustbox}

\end{table}

\subsection{Training and Recognition Protocols}
\label{sect:protocol}

The networks are trained for ocular identification using cross-entropy loss on VGG2 crops.
We use SGDM (batch=128, learning rate=0.01, 0.005, 0.001, 0.0001, decreased when the validation loss plateaus) and set aside 2\% of training images per user for validation.
The models are initialized from scratch, ImageNet, or face recognition weights (Section~\ref{sect:database}).
For scratch/ImageNet, the classification head is adjusted to 8631 classes, whereas with face-pretraining, it remains unchanged.

Verification is performed on the 368 users of VGGFace2-Pose, both intra- and cross-pose. 
Identity templates per user are created by extracting the descriptors of the left and right eyes from the layer adjacent to the classification layer (i.e., the Global Average Pooling).
Given a pair of face images, the left and right eyes are compared separately, and the two scores are averaged.
As comparison metrics, we use the cosine similarity and the $\chi^2$ distance.
Cosine is standard in CNN-based verification, but $\chi^2$ has also shown good performance \cite{[Hernandez18]}.
Genuine scores are obtained by comparing the eye crops of one face image against the rest of the same user (excluding symmetric matches)
For impostor scores, the crops of the 1$^{st}$ face image of a user are compared with the 2$^{nd}$ image of the remaining users. 
For UFPR, we follow its predefined protocol of three folds, testing on 374 users per fold. 
As with VGGFace2-Pose, eye crops are compared separately and scores averaged.
%
%Genuine comparisons are done in the same way as VGG2 by comparing all against all face images of a user.
%
Table~\ref{tab:scores} summarizes the number of score comparisons.

\begin{figure}[t]
\centering
\includegraphics[width=0.88\textwidth]{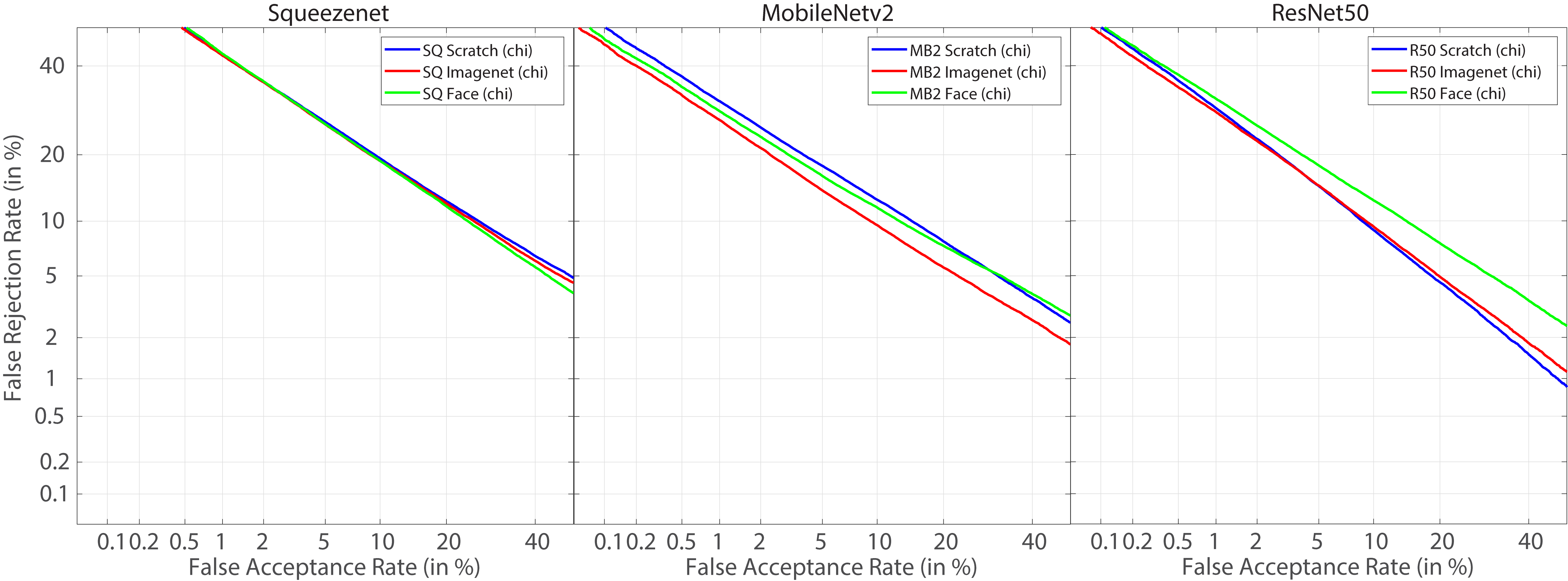}
\caption{Ocular verification results on VGGFace2-Pose for different network initializations ($\chi^2$ distance).} \label{fig:det_vgg2_pose}
\end{figure}

\begin{figure}[t]
\centering
\includegraphics[width=0.44\textwidth]{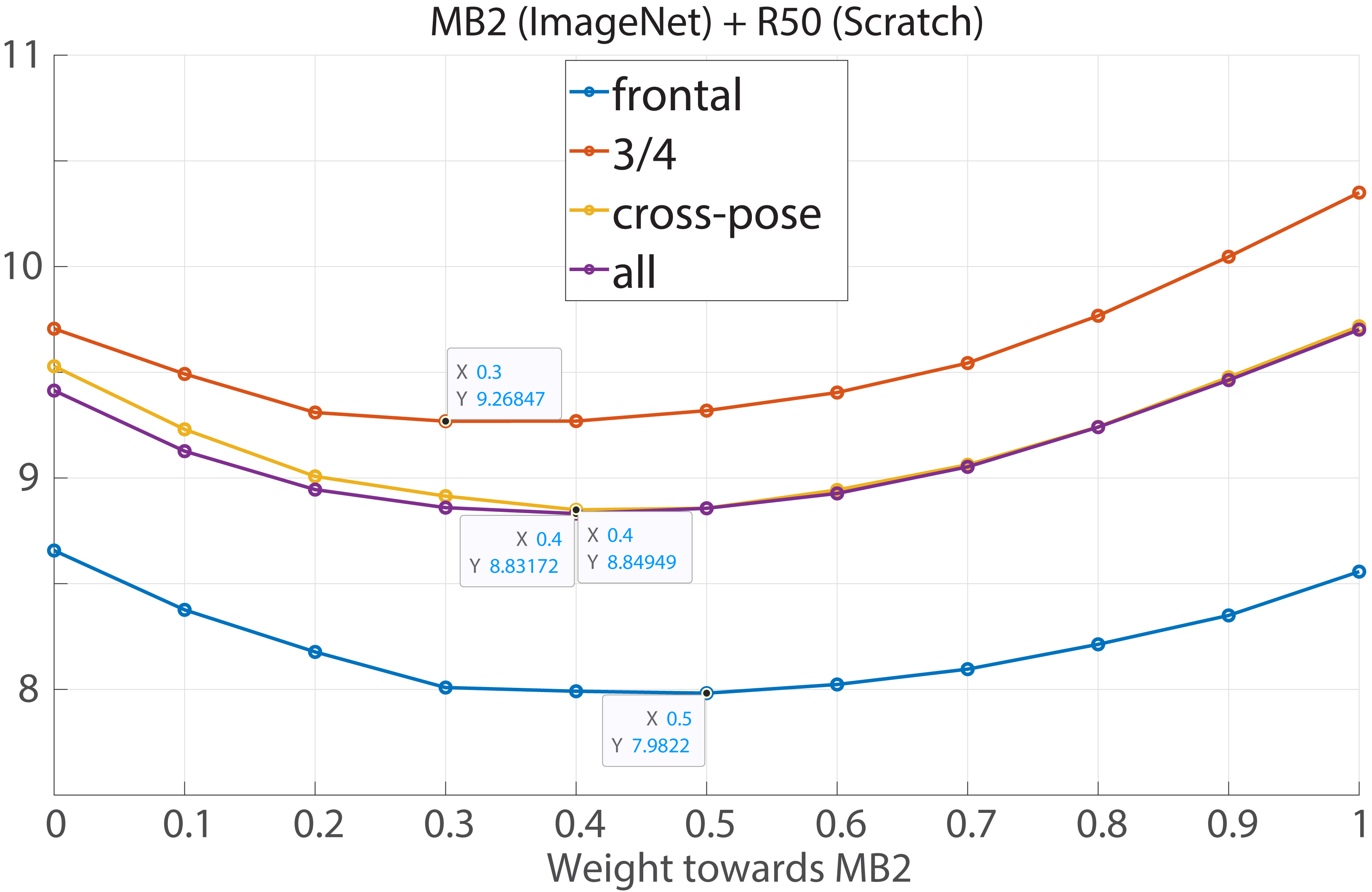}
\includegraphics[width=0.44\textwidth]{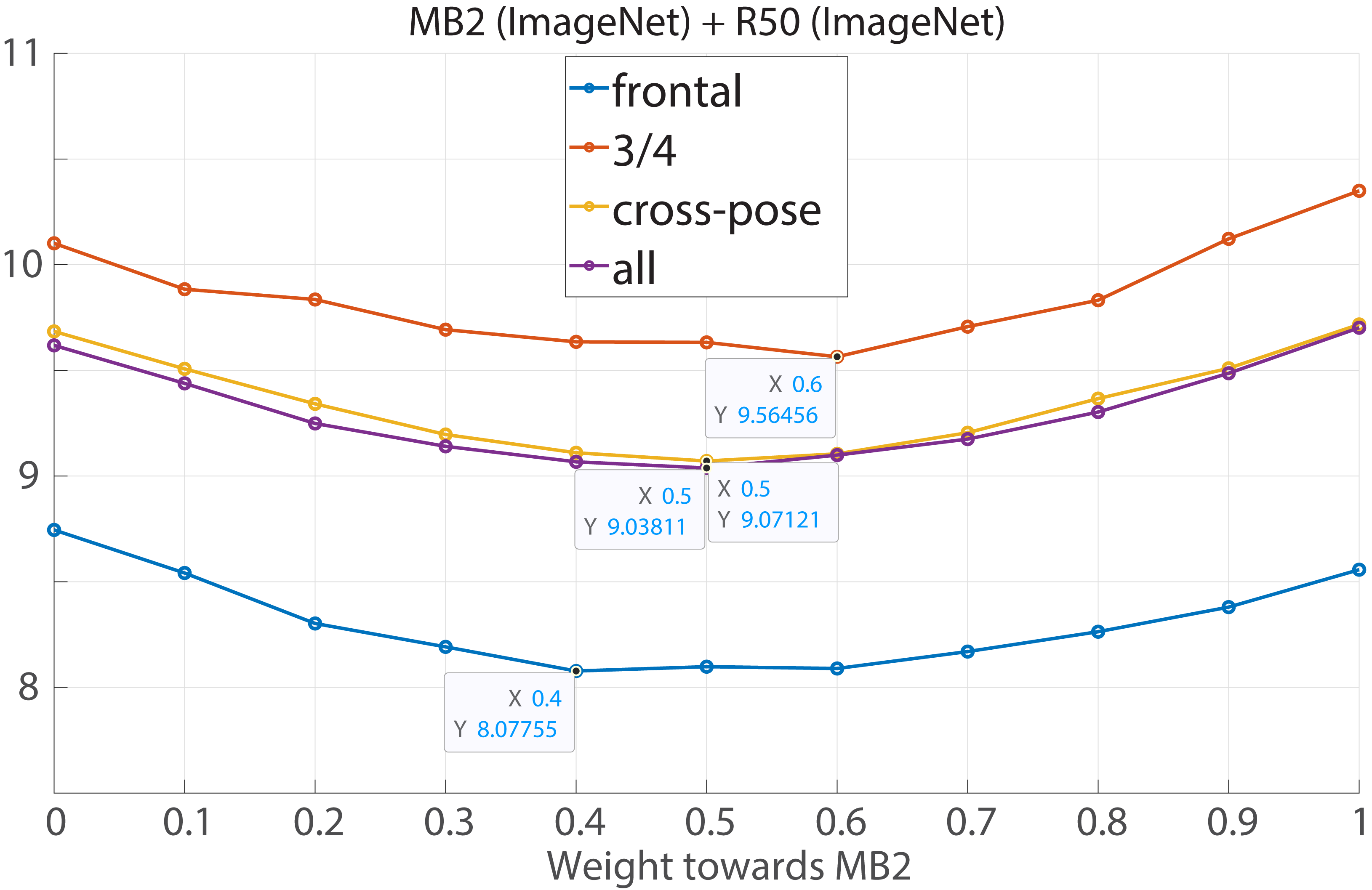}
\caption{Ocular fusion verification results on VGGFace2-Pose for different network initializations (EER \%, $\chi^2$ distance). Left: best initialization per network according to Table~\ref{tab:eer_vgg2_pose}. Right: all networks initialized on ImageNet.} \label{fig:eer_vgg2_pose_MB2_R50_fusion}
\end{figure}

\section{Results with VGGFace2-Pose database}
\label{sect:results_VGG2}

%\subsection{Experiments with VGGFace2-Pose}

We first report (Table~\ref{tab:eer_vgg2_pose}) ocular verification results of the networks on VGGFace2-Pose.
A first observation is that $\chi^2$ distance (right part of the table) consistently provides better results than the cosine similarity.
This is in consonance with previous works \cite{[Hernandez18]}.
In some cases, the difference in favour of $\chi^2$ is more than 1\% of EER reduction.
%
%While cosine measures the angle between the embedding vectors, $\chi^2$  encodes local relative differences between channels, which may be more suitable for low-quality images such as VGG2.
%
Regarding network initialisation, there is no consensus on the best strategy (bold numbers). For SqueezeNet, face recognition initialisation works best; for MobileNetv2, the best results are achieved with ImageNet initialisation; and for ResNet50, training from scratch is the best case. 
It may be intuitive to assume that fine-tuning face recognition networks for the ocular modality would be the best option, since the network had already \textit{seen} eye regions. However, this is seen to be detrimental. One possible explanation would be that face networks may be too specialised already for the full-face. On the other hand, ImageNet or scratch initialisation allows to start with more primitive features (edges, corners, etc.) that adapt better to the ocular task.
%
%Even networks with just ImageNet or, in some cases, with random weights and no further training have been shown to yield surprisingly good ocular performance in previous studies \cite{Hernandez23access_oneshot}.
%
By looking at the DET curves (Figure~\ref{fig:det_vgg2_pose}, $\chi^2$ only), conclusions about the best initialisation derived from the EER also hold, with the only exception that ResNet50 works better with ImageNet start in some regions of the DET.
This also suggests that ImageNet pretraining can offer good robustness and serve as a general starting point for specialised features, as seen in countless works in the computer vision literature \cite{[Razavian14]}, and not just in biometrics \cite{[Alonso22inffus],Alonso24wifs_cnn_vit_ots,[Hernandez18],[Hernandez19],Hernandez23access_oneshot,[Nguyen18]}. 
\begin{table}[t]
\centering
\caption{Ocular verification results on UFPR for different network initializations (EER/AUC \% with OW/CW protocol). The best result of each network (per column) is marked in bold. The table also shows ocular recognition results from previous works on the same database and protocol.}
\label{tab:eer_AUC_UFPR}

\resizebox{0.85\textwidth}{!}{%

\begin{tabular}{llllllllll}
 &  &  &  &  &  & & & & \\ \cline{3-10} 
 
 & \multicolumn{1}{l||}{} & \multicolumn{4}{c||}{\textbf{cosine similarity}} & \multicolumn{4}{c||}{\textbf{$\chi^2$ distance}} \\ \cline{3-10} 
 
 & \multicolumn{1}{l||}{} & \multicolumn{2}{c||}{\textbf{EER}} & \multicolumn{2}{c||}{\textbf{AUC}} & \multicolumn{2}{c||}{\textbf{EER}} & \multicolumn{2}{c||}{\textbf{AUC}}  \\ \hline
 
\multicolumn{1}{|l|}{\textbf{Net}} & \multicolumn{1}{l||}{\textbf{Initialization}} & \multicolumn{1}{l|}{\textbf{avg}} & \multicolumn{1}{l||}{\textbf{std}} & \multicolumn{1}{l|}{\textbf{avg}} & \multicolumn{1}{l||}{\textbf{std}} & \multicolumn{1}{l|}{\textbf{avg}} & \multicolumn{1}{l||}{\textbf{std}} & \multicolumn{1}{l|}{\textbf{avg}} & \multicolumn{1}{l||}{\textbf{std}} \\ \hline \hline

\multicolumn{1}{|l|}{\multirow{3}{*}{SQ}} & \multicolumn{1}{l||}{Scratch} & \multicolumn{1}{l|}{2.57} & \multicolumn{1}{l||}{\textbf{0.17}} & \multicolumn{1}{l|}{99.59} & \multicolumn{1}{l||}{0.10} & \multicolumn{1}{l|}{2.47} & \multicolumn{1}{l||}{0.18} & \multicolumn{1}{l|}{99.62} & \multicolumn{1}{l||}{\textbf{0.08}} \\ \cline{2-10} 

\multicolumn{1}{|l|}{} & \multicolumn{1}{l||}{ImageNet} & \multicolumn{1}{l|}{\textbf{2.13}} & \multicolumn{1}{l||}{0.18} & \multicolumn{1}{l|}{\textbf{99.69}} & \multicolumn{1}{l||}{\textbf{0.08}} & \multicolumn{1}{l|}{\textbf{2.07}} & \multicolumn{1}{l||}{\textbf{0.13}} & \multicolumn{1}{l|}{\textbf{99.70}} & \multicolumn{1}{l||}{\textbf{0.08}} \\ \cline{2-10} 

\multicolumn{1}{|l|}{} & \multicolumn{1}{l||}{Face} & \multicolumn{1}{l|}{3.05} & \multicolumn{1}{l||}{0.30} & \multicolumn{1}{l|}{99.50} & \multicolumn{1}{l||}{0.13} & \multicolumn{1}{l|}{2.80} & \multicolumn{1}{l||}{0.25} & \multicolumn{1}{l|}{99.55} & \multicolumn{1}{l||}{0.11} \\ \hline \hline

\multicolumn{1}{|l|}{\multirow{3}{*}{MB2}} & \multicolumn{1}{l||}{Scratch} & \multicolumn{1}{l|}{1.98} & \multicolumn{1}{l||}{0.12} & \multicolumn{1}{l|}{99.76} & \multicolumn{1}{l||}{0.05} & \multicolumn{1}{l|}{1.92} & \multicolumn{1}{l||}{0.11} & \multicolumn{1}{l|}{99.77} & \multicolumn{1}{l||}{0.04} \\ \cline{2-10} 

\multicolumn{1}{|l|}{} & \multicolumn{1}{l||}{ImageNet} & \multicolumn{1}{l|}{\textbf{1.53}} & \multicolumn{1}{l||}{\textbf{0.08}} & \multicolumn{1}{l|}{\textbf{99.85}} & \multicolumn{1}{l||}{\textbf{0.02}} & \multicolumn{1}{l|}{\textbf{1.49}} & \multicolumn{1}{l||}{0.09} & \multicolumn{1}{l|}{\textbf{99.86}} & \multicolumn{1}{l||}{\textbf{0.02}}  \\ \cline{2-10} 

\multicolumn{1}{|l|}{} & \multicolumn{1}{l||}{Face} & \multicolumn{1}{l|}{1.98} & \multicolumn{1}{l||}{\textbf{0.08}} & \multicolumn{1}{l|}{99.75} & \multicolumn{1}{l||}{0.03} & \multicolumn{1}{l|}{1.90} & \multicolumn{1}{l||}{\textbf{0.06}} & \multicolumn{1}{l|}{99.76} & \multicolumn{1}{l||}{0.03} \\ \hline \hline

\multicolumn{1}{|l|}{\multirow{3}{*}{R50}} & \multicolumn{1}{l||}{Scratch} & \multicolumn{1}{l|}{2.00} & \multicolumn{1}{l||}{0.11} & \multicolumn{1}{l|}{99.78} & \multicolumn{1}{l||}{0.03} & \multicolumn{1}{l|}{1.90} & \multicolumn{1}{l||}{0.08} & \multicolumn{1}{l|}{99.80} & \multicolumn{1}{l||}{0.03} \\ \cline{2-10} 

\multicolumn{1}{|l|}{} & \multicolumn{1}{l||}{ImageNet} & \multicolumn{1}{l|}{\textbf{1.47}} & \multicolumn{1}{l||}{\textbf{0.04}} & \multicolumn{1}{l|}{\textbf{99.87}} & \multicolumn{1}{l||}{\textbf{0.01}}  & \multicolumn{1}{l|}{\textbf{1.41}} & \multicolumn{1}{l||}{\textbf{0.03}} & \multicolumn{1}{l|}{\textbf{99.88}} & \multicolumn{1}{l||}{\textbf{0.01}} \\ \cline{2-10} 

\multicolumn{1}{|l|}{} & \multicolumn{1}{l||}{Face} & \multicolumn{1}{l|}{1.94} & \multicolumn{1}{l||}{0.08} & \multicolumn{1}{l|}{99.78} & \multicolumn{1}{l||}{0.03} & \multicolumn{1}{l|}{1.85} & \multicolumn{1}{l||}{0.04} & \multicolumn{1}{l|}{99.79} & \multicolumn{1}{l||}{0.02} \\ \hline \hline

\multicolumn{2}{|l|}{MB2+R50 (ImageNet)} & \multicolumn{1}{l|}{-} & \multicolumn{1}{l||}{-} & \multicolumn{1}{l|}{-} & \multicolumn{1}{l||}{-} & \multicolumn{1}{l|}{\textbf{1.27}} & \multicolumn{1}{l||}{\textbf{0.03}} & \multicolumn{1}{l|}{-} & \multicolumn{1}{l||}{-}  \\ \hline 

\end{tabular}

}

\end{table}

\begin{figure}[t]
\centering
\includegraphics[width=0.88\textwidth]{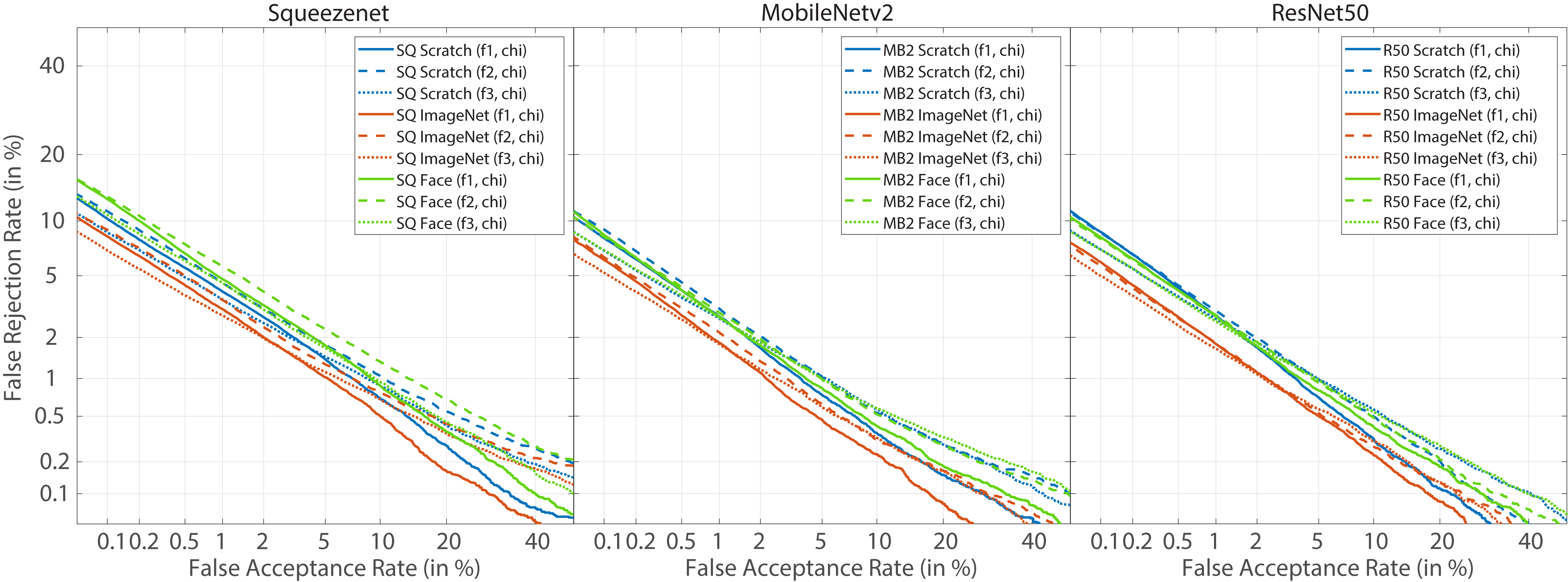}
\caption{Ocular verification on UFPR test folds for different network initializations ($\chi^2$ distance).} \label{fig:det_ufpr}
\end{figure}

In terms of absolute performance, the residual-based networks MobileNetv2 and ResNet50 provide much better EER than the simpler SqueezeNet. 
It can also be seen that the comparison of ocular images extracted from frontal face images is slightly better than three-quarter or cross-pose comparisons (more than 1\% difference in EER with ResNet50 and $\sim$2\% or more with the other networks). This can be expected given the progressive distortion of the ocular region as the view departs from frontal.
The bottom part of Table~\ref{tab:eer_vgg2_pose} also shows the face recognition performance reported on previous works with the same database.
As it can be seen, performance with the full face on VGGFace2-Pose is significantly better than ocular (EER in the range of 3-6\% vs 9-15\%).
We attribute this to the quality of VGG2 data \cite{[Cao18vggface2]}, which consist of images with significant variability in pose, illumination, etc., providing richer information when the entire face is visible. 
In contrast, ocular crops represent a zoomed, more limited region with less discriminative content under such conditions.

The two best performing networks, MobileNetv2 and ResNet50, were observed previously to be highly complementary for face recognition via score fusion \cite{Alonso23wifs_lime_biometrics}.
We also assess here their complementarity for ocular recognition (Figure~\ref{fig:eer_vgg2_pose_MB2_R50_fusion}).
This is done by combining their verification scores, denoted as $s_{MB2}$ and $s_{R50}$, through a weighted average approach via $a \times s_{MB2} + (a-1) \times s_{R50}$ %(MB2=MobileNetv2, R50=ResNet50)
($a\in [0,1]$).
Figure~\ref{fig:eer_vgg2_pose_MB2_R50_fusion} shows the results for different values of the weight $a$ (support towards MobileNetv2).
We test two cases: the best initialization per network according to Table~\ref{tab:eer_vgg2_pose}, and all networks initialized on ImageNet, since we observed above that ImageNet is a good general starting point. 
We also tested other fusion combinations involving SqueezeNet, but they did not provide any performance gain due to the much worse individual performance of such network, so results of those experiments are omitted.

Notably, the fusion of MobileNetv2 and ResNet50 enhances performance, with the optimal achieved when both networks are assigned a roughly equal weight ($a$ between 0.4 and 0.6). 
We select the cases with the highest overall accuracy ($a=0.4$ for the best initialization and $a=0.5$ for ImageNet initialization) and provide its exact EER values in Table~\ref{tab:eer_vgg2_pose}. 
Using the best initialization per network provides a slight advantage compared to ImageNet initialization in both networks (overall EER of 8.83\% vs 9.04\%).
In addition, the fusion enhances performance across all pose cases, providing EER gains of more than 0.4\%.

\section{Results with UFPR database}
\label{sect:results_UFPR}

%\subsection{Experiments with UFPR}

We then evaluate the ocular recognition networks trained with VGG2 on the UFPR-Periocular dataset \cite{zanlorensi2022_SR_UFPR_db}.
Results are given in Table~\ref{tab:eer_AUC_UFPR}.
We follow the OW/CW protocol and reporting metrics of the UFPR paper (EER and AUC) across the three test folds.
A first evident observation is the lower EERs in comparison to VGGFace2-Pose (Table~\ref{tab:eer_vgg2_pose}).
UFPR is a purposely-captured ocular database, with users employing their mobiles in selfie mode while looking frontally to the device. In principle, this provides higher resolution and quality ocular images in a more controlled setup, since users are asked to place their eyes in a region of interest shown in the device screen.
In contrast, VGG2 images are face images captured in-the-wild, of which we crop the smaller ocular area.
As as result, the EERs with UFPR are in the range of 1-2\% (even with SqueezeNet), compared to 9-15\% with VGGFace2-Pose.
With UFPR, it can also be observed an advantage in favour of the $\chi^2$ distance, although in this case the differences are in general less than 0.1\%.
While cosine measures the angle between the embedding vectors, $\chi^2$ encodes local relative differences between channels, which may be more suitable for low-quality images such as VGG2, where higher EER gains were observed by using $\chi^2$.
Regarding initialization, ImageNet wins in all cases with UFPR by a large margin. 
The DET curves (Figure~\ref{fig:det_ufpr}, $\chi^2$ only) support this conclusion, i.e. the red curves (ImageNet initialization) win in nearly all regions, with the exception of SqueezeNet, where at low FRR, other initialization are seen to work better. 
In any case, this confirms our above observations that ImageNet pretraining constitute a good starting point overall, even if we have a large amount of training images.

\begin{figure}[t]
\centering
\includegraphics[width=0.88\textwidth]{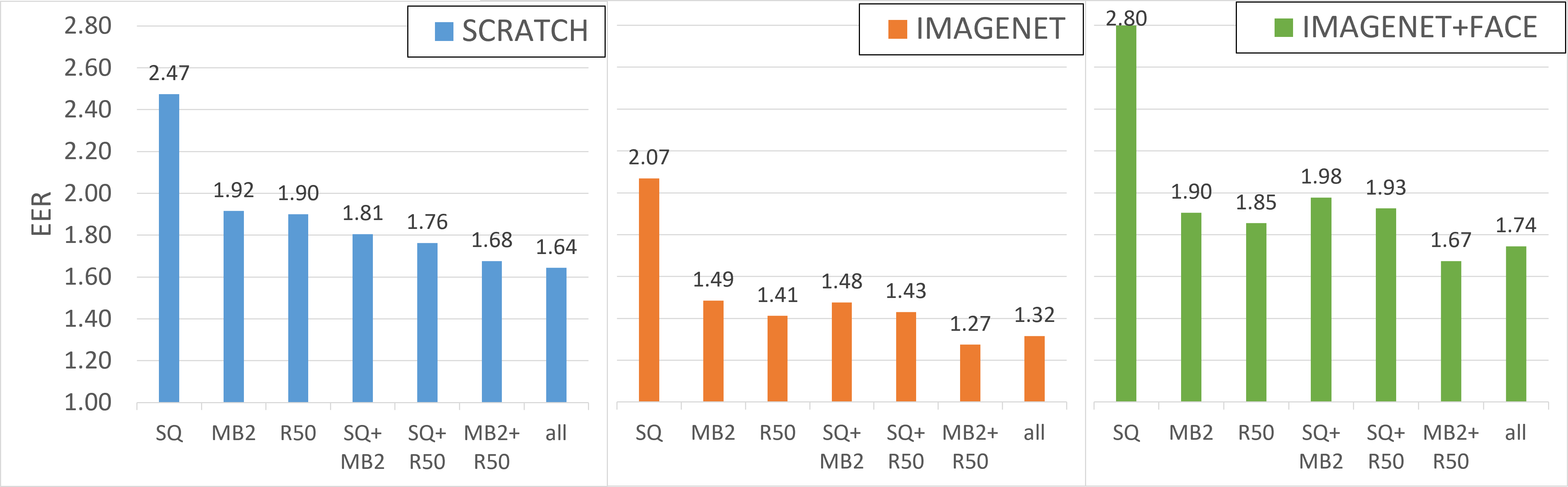}
\caption{Ocular fusion results on UFPR for different network initializations ($\chi^2$ distance, average EER\% of the folds, equal weight per network in the fusion).} \label{fig:eer_ufpr_fusion}
\end{figure}

\begin{figure}[t]
\centering
\includegraphics[width=0.44\textwidth]{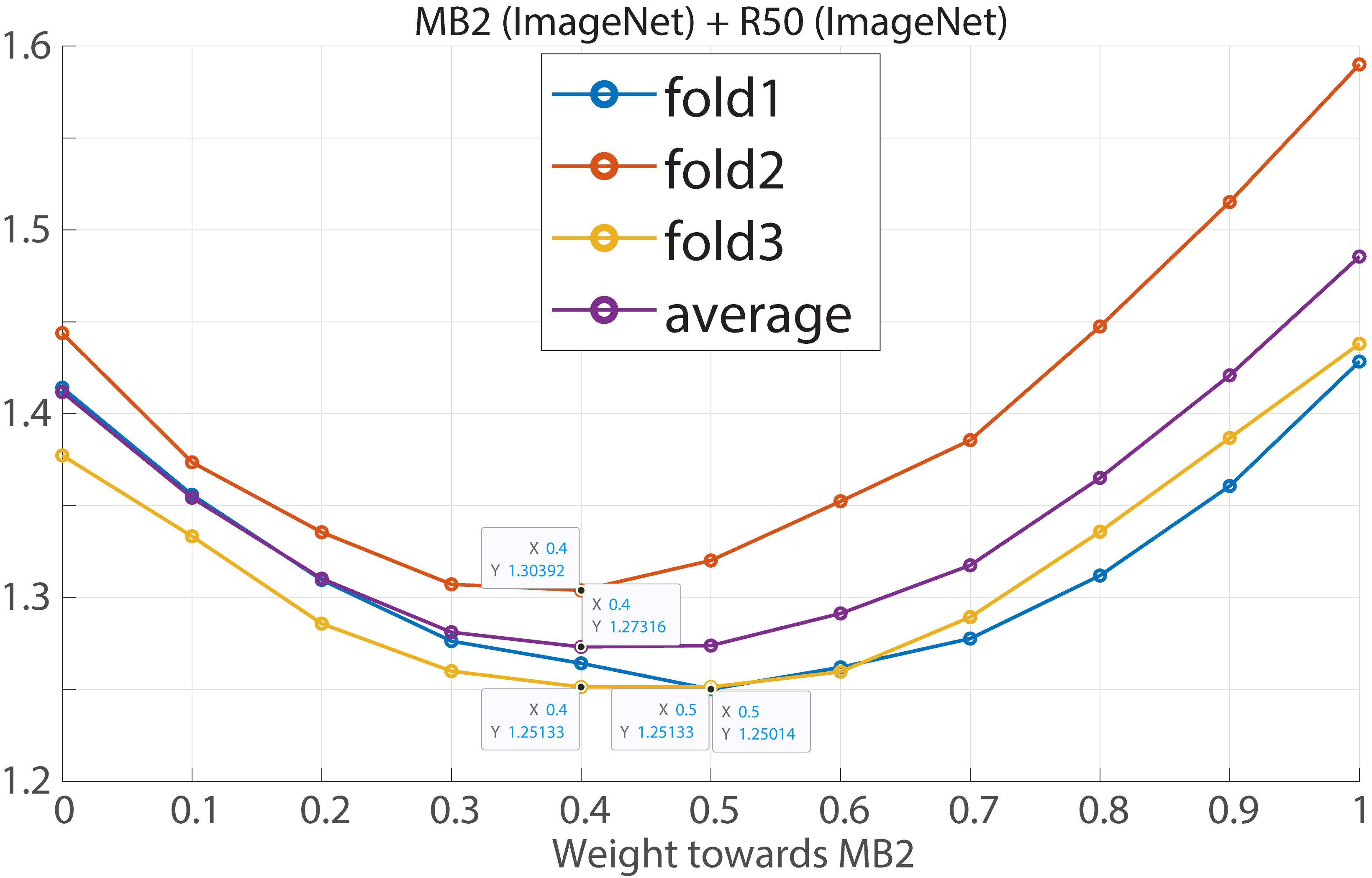}
\caption{Ocular fusion verification on UFPR for networks initialized on ImageNet (EER \%, $\chi^2$ distance).} \label{fig:eer_ufpr_MB2_R50_fusion}
\end{figure}

As in the previous sub-section, we also analyze network complementarity.
Figure~\ref{fig:eer_ufpr_fusion} shows results of average score fusion combination, with each network given the same weight in the fusion.
In this case, it can be seen that even involving SqueezeNet in the fusion provides performance gains if the networks are initialized from scratch (left). However, this is not the initialization providing the best absolute EERs. With the other two initializations, involving SqueezeNet does not provide any fusion benefit.
We further show (Figure~\ref{fig:eer_ufpr_MB2_R50_fusion}) the combination of MobileNetv2 and ResNet50 for different supports towards each network in the weighted fusion. 
We only show the case where both networks are initialized with ImageNet, since this was the best case for both networks with UFPR. 
Again, the fusion is seen to improve performance, being optimal when both networks receive approximately the same weight ($a$ between 0.4-0.5).
We select $a$=0.4 as the case with the best average EER and provide the exact values in Table~\ref{tab:eer_AUC_UFPR}, where it can be seen that this is the best case overall.

We finally compare (Table~\ref{tab:eer_AUC_UFPR_other_works}) our results with previous works employing the same OW/CW evaluation protocol on the UFPR database. 
Works such as \cite{Rattani23ACCESS_OcularCNNPruningBenchmark,Coelho24_lacci_PeriocularEfficientNet} are deliberately left out, since they employed a different training/testing protocol. %, so they are not directly comparable.
The work \cite{zanlorensi2022_SR_UFPR_db} corresponds to the seminal paper of UFPR, which established the baseline performance, improved later on by more recent research \cite{Kolf22ijcb_light_ocular_lowbit_quantization,Kolf23_ivc_syper_ocular_db,Kolf24_eaai_MixQuantBio_face_ocular}.
However, such works made use of UFPR as the unique training database, with just contains 33,660 images. 
This is surpassed by our training strategy, consisting of 1,907,572 ocular crops from the large VGGFace2 database, which provides state-of-the-art performance with the UFPR database.

% \usepackage{multirow}
\begin{table}[t]
\centering
\caption{Ocular verification results on UFPR from previous works (EER/AUC \% with the OW/CW protocol). Works with (*) flipped left/right eyes to the same orientation and considered both eyes to have the same identity.}
\label{tab:eer_AUC_UFPR_other_works}

\resizebox{0.85\textwidth}{!}{%

\begin{tabular}{llllll}
 &  &  &  &  &  \\ \cline{3-6} 

 \multicolumn{2}{l||}{} & \multicolumn{2}{c||}{\textbf{EER}} & \multicolumn{2}{c||}{\textbf{AUC}} \\ \hline
 
\multicolumn{1}{|l||}{\textbf{Net}}  & \multicolumn{1}{c||}{\textbf{Training}} & \multicolumn{1}{l|}{\textbf{avg}} & \multicolumn{1}{l||}{\textbf{std}} & \multicolumn{1}{l|}{\textbf{avg}} & \multicolumn{1}{l||}{\textbf{std}} \\ \hline \hline

\multicolumn{1}{|l||}{MobileNetv2 \cite{zanlorensi2022_SR_UFPR_db}} & \multicolumn{1}{c||}{ImageNet +} & \multicolumn{1}{c|}{3.17} & \multicolumn{1}{c||}{0.33} & \multicolumn{1}{c|}{99.56} & \multicolumn{1}{c||}{0.08} \\ 

\multicolumn{1}{|l||}{DenseNet121 \cite{zanlorensi2022_SR_UFPR_db}} & \multicolumn{1}{c||}{UFPR} & \multicolumn{1}{c|}{3.39} & \multicolumn{1}{c||}{0.46}  & \multicolumn{1}{c|}{99.51} & \multicolumn{1}{c||}{0.12} \\ 

\multicolumn{1}{|l||}{ResNet50 \cite{zanlorensi2022_SR_UFPR_db}} & \multicolumn{1}{c||}{} & \multicolumn{1}{c|}{5.98} & \multicolumn{1}{c||}{0.67} & \multicolumn{1}{c|}{98.60} & \multicolumn{1}{c||}{0.28} \\ 

\multicolumn{1}{|l||}{VGG16 \cite{zanlorensi2022_SR_UFPR_db}} & \multicolumn{1}{c||}{} & \multicolumn{1}{c|}{8.52} & \multicolumn{1}{c||}{0.92} & \multicolumn{1}{c|}{97.38} & \multicolumn{1}{c||}{0.53} \\ \hline 

\multicolumn{1}{|l||}{ResNet50-Face \cite{zanlorensi2022_SR_UFPR_db}} & \multicolumn{1}{c||}{VGGFace1 +} & \multicolumn{1}{c||}{4.38} & \multicolumn{1}{c||}{0.47} & \multicolumn{1}{c|}{99.18} & \multicolumn{1}{c||}{0.16} \\ 

\multicolumn{1}{|l||}{VGG16-Face \cite{zanlorensi2022_SR_UFPR_db}} & \multicolumn{1}{c||}{UFPR} & \multicolumn{1}{c|}{7.78} & \multicolumn{1}{c||}{0.75} & \multicolumn{1}{c|}{97.70} & \multicolumn{1}{c||}{0.42} \\ \hline \hline

\multicolumn{1}{|l||}{ResNet18 W8A8 \cite{Kolf22ijcb_light_ocular_lowbit_quantization} *} & \multicolumn{1}{c||}{UFPR} & \multicolumn{1}{c|}{5.99} & \multicolumn{1}{c||}{0.39} & \multicolumn{1}{c|}{98.41} & \multicolumn{1}{c||}{0.16} \\

\multicolumn{1}{|l||}{ResNet50 W8A8 \cite{Kolf22ijcb_light_ocular_lowbit_quantization} *} & \multicolumn{1}{c||}{(scratch)} & \multicolumn{1}{c|}{5.99} & \multicolumn{1}{c||}{0.41} & \multicolumn{1}{c|}{98.39} & \multicolumn{1}{c||}{0.18} \\ 

\multicolumn{1}{|l||}{MobileFaceNet W6A6 \cite{Kolf22ijcb_light_ocular_lowbit_quantization} *} & \multicolumn{1}{c||}{} & \multicolumn{1}{c|}{4.02} & \multicolumn{1}{c||}{0.19} & \multicolumn{1}{c|}{99.18} & \multicolumn{1}{c||}{0.06} \\ \hline \hline

\multicolumn{1}{|l||}{ResNet18 MQ \cite{Kolf24_eaai_MixQuantBio_face_ocular} *} & \multicolumn{1}{c||}{UFPR} & \multicolumn{1}{c|}{2.63} & \multicolumn{1}{c||}{-} & \multicolumn{1}{c|}{99.60}  & \multicolumn{1}{c||}{-} \\ 

\multicolumn{1}{|l||}{ResNet50 MQ \cite{Kolf24_eaai_MixQuantBio_face_ocular} *} & \multicolumn{1}{c||}{(scratch)} & \multicolumn{1}{c|}{2.59} & \multicolumn{1}{c||}{-} & \multicolumn{1}{c|}{99.61}  & \multicolumn{1}{c||}{-} \\ 

\multicolumn{1}{|l||}{MobileFaceNet MQ \cite{Kolf24_eaai_MixQuantBio_face_ocular} *} & \multicolumn{1}{c||}{} & \multicolumn{1}{c|}{2.80} & \multicolumn{1}{c||}{-} & \multicolumn{1}{c|}{99.55}  & \multicolumn{1}{c||}{-} \\ \hline \hline

\multicolumn{1}{|l||}{ResNet18 FP32 \cite{Kolf22ijcb_light_ocular_lowbit_quantization,Kolf23_ivc_syper_ocular_db,Kolf24_eaai_MixQuantBio_face_ocular} *} & \multicolumn{1}{c||}{UFPR} & \multicolumn{1}{c|}{5.76} & \multicolumn{1}{c||}{0.38} & \multicolumn{1}{c|}{98.51} & \multicolumn{1}{c||}{0.15} \\ 

\multicolumn{1}{|l||}{ResNet50 FP32 \cite{Kolf22ijcb_light_ocular_lowbit_quantization,Kolf23_ivc_syper_ocular_db,Kolf24_eaai_MixQuantBio_face_ocular} *} & \multicolumn{1}{c||}{(scratch)} & \multicolumn{1}{c|}{5.88} & \multicolumn{1}{c||}{0.38} & \multicolumn{1}{c|}{98.47} & \multicolumn{1}{c||}{0.17} \\

\multicolumn{1}{|l||}{MobileFaceNet FP32 \cite{Kolf22ijcb_light_ocular_lowbit_quantization,Kolf23_ivc_syper_ocular_db,Kolf24_eaai_MixQuantBio_face_ocular} *} & \multicolumn{1}{c||}{} & \multicolumn{1}{c|}{3.86} & \multicolumn{1}{c||}{0.21} & \multicolumn{1}{c|}{99.23} & \multicolumn{1}{c||}{0.05} \\ \hline \hline

\multicolumn{1}{|l||}{SqueezeNet (this work) *} & \multicolumn{1}{c||}{ImageNet + } & \multicolumn{1}{c|}{2.07} & \multicolumn{1}{c||}{0.13} & \multicolumn{1}{c|}{99.70}  & \multicolumn{1}{c||}{0.08} \\

\multicolumn{1}{|l||}{MobileNetv2  (this work) *} & \multicolumn{1}{c||}{VGG2 ocular} & \multicolumn{1}{c|}{1.49} & \multicolumn{1}{c||}{0.09} & \multicolumn{1}{c|}{99.86}  & \multicolumn{1}{c||}{0.02} \\ 

\multicolumn{1}{|l||}{ResNet50 (this work) *} & \multicolumn{1}{c||}{} & \multicolumn{1}{c|}{\textbf{1.41}} & \multicolumn{1}{c||}{\textbf{0.03}} & \multicolumn{1}{c|}{\textbf{99.88}}  & \multicolumn{1}{c||}{\textbf{0.01}} \\ 

\multicolumn{1}{|l||}{MB2+R50 (this work) *} & \multicolumn{1}{c||}{} & \multicolumn{1}{c|}{\textbf{1.27}} & \multicolumn{1}{c||}{\textbf{0.03}} & \multicolumn{1}{c|}{-}  & \multicolumn{1}{c||}{-} \\ \hline

\end{tabular}

}

\end{table}

\section{Conclusions}

We address the task of developing biometric deep-recognition models that employ periocular images. 
%
%Partial faces are an issue in both unconstrained and controlled setups, like masks, social media \cite{Hedman22_pr_selfie_beauty_filters}, pollution, cultural coverage, work gear, etc. \cite{sharma23cviu_periocular_masks_survey}.
%
%This region, which surrounds the eye, remains available at various distances, even with partial face occlusion at close distances, or low iris resolution due to long distances \cite{Alonso24computers_periSOA}.
%
%This makes the periocular modality a suitable alternative, with minimal cooperation compared to face or iris, and availability at a wider range of distances when face may appear occluded or iris may not have sufficient resolution to even be properly segmented \cite{Alonso24computers_periSOA}.
%
%
In this work, we have evaluated three architectures of varying complexity (SqueezeNet, MobileNetv2, and ResNet50) trained on 1,907,572 periocular crops extracted from the large-scale VGGFace2 (VGG2) dataset \cite{[Cao18vggface2]}, after filtering out non-frontal and very low-resolution images.
This contrasts with prevalent research \cite{zanlorensi22_AIR_ocular_db_competitions_survey}, including recent works \cite{zanlorensi2022_SR_UFPR_db,Kolf22ijcb_light_ocular_lowbit_quantization,Kolf23_ivc_syper_ocular_db,Rattani23ACCESS_OcularCNNPruningBenchmark,Coelho24_lacci_PeriocularEfficientNet,Kolf24_eaai_MixQuantBio_face_ocular}, which rely on small-scale periocular databases with only a few thousand images.

We test multiple initialization strategies of the networks, including scratch, ImageNet weights, and fine-tuning of face recognition models trained on VGG2. 
%
%The latter approach takes advantage of the availability of pre-trained face recognition networks using the original large-scale VGGFace2 database.
%
Intuitively, fine-tuning face recognition networks for the ocular
modality would be the best option, since face images already contain the ocular region.
However, we observed that ImageNet weights are a better general starting point, whereas fine-tuning a face network is actually detrimental. 
We hypothesize that a face network may be already too specialized, whereas a more primitive initialization like ImageNet allows the networks to adapt better to the ocular images.
Even networks with just ImageNet or, in some cases, with random weights and no further training have been shown to yield surprisingly good ocular performance in previous studies \cite{Hernandez23access_oneshot}.

We carry out our evaluation experiments with two sets, the VGGFace2-Pose, a subset of VGG2 \cite{[Cao18vggface2]}, and the UFPR-Periocular database \cite{zanlorensi2022_SR_UFPR_db}.
Since VGG2 images are captured in-the-wild and we employ ocular crops of already low-quality face images, the EERs are modest with VGGFace2-Pose (9-15\%), compared to 3-6\% with full-face input.
In contrast, the more controlled acquisition of UFPR selfies leads to EERs of 1-2\% which, to our knowledge, are the best reported results on this dataset
%
%This validates our approach of employing ocular crops from a large-scale face database, in contrast with previous studies which employ smaller ocular database for training.
%
In addition, two of the employed networks (MobileNetv2 and ResNet50) are found to be complementary, observing a performance improvement by just combining (averaging) their decision scores.

We expect to achieve further gains by incorporating margin-based losses such as ArcFace, already employed in some works that we surpassed \cite{Kolf22ijcb_light_ocular_lowbit_quantization,Kolf23_ivc_syper_ocular_db,Kolf24_eaai_MixQuantBio_face_ocular}.
We also hypothesize that a sequential fine-tuning with ocular crops from the MS-Celeb-1M (MS1M) \cite{[Guo16_MSCeleb1M]} and VGGFace2 \cite{[Cao18vggface2]} databases would provide even more benefit. 
This approach was followed earlier for face recognition, providing superior performance compared to just using VGGFace2 \cite{[Cao18vggface2],[Alonso20SqueezeFacePoseNet]}, since MS1M has more images overall, but VGGFace2 has more intra-class diversity.

\section*{Acknowledgements} 

This work was partly done while F. A.-F. was a visiting researcher at the University of the Balearic Islands.
F. A.-F., K. H.-D., and J. B. thank the
Swedish Research Council (VR) and the EU (HORIZON Europe project PopEye under Grant Agreement no 101168317) for funding their research.
Funded by the European Union. Views and opinions expressed are, however, those of the author(s) only and do not necessarily reflect those of the European Union or the European Research Executive Agency. Neither the European Union nor the granting authority can be held responsible for them.
This work is part of the Project PID2022-136779OB-C32 (PLEISAR) funded by MICIU/ AEI /10.13039/501100011033/ and FEDER, EU.
%
%Author J. M. B. thanks the project EXPLAINING - "Project EXPLainable Artificial INtelligence systems for health and well-beING", under Spanish national projects funding (PID2019-104829RA-I00/AEI/10.13039/501100011033).

{\small

\bibliographystyle{splncs04}

%\bibliography{fernando1}

}

\end{document}